\documentclass[10pt,journal,compsoc]{IEEEtran}
\usepackage{amsmath,amsfonts}
\usepackage{algorithmic}
\usepackage{algorithm}
\usepackage{array}
\usepackage[caption=false,font=footnotesize,labelfont=rm,textfont=rm]{subfig}
\usepackage{textcomp}
\usepackage{stfloats}
\usepackage{balance}
\usepackage{url}
\usepackage{verbatim}
\usepackage{graphicx}
\usepackage{cite}
\usepackage{multirow}
\usepackage[numbers]{natbib}
\ifCLASSINFOpdf
\else
\fi
\begin{document}
%
\title{Emulating Reader Behaviors for Fake News Detection}
%
%
%
%

\author{Junwei Yin,
        Min Gao, Kai Shu, Zehua Zhao, Yinqiu Huang,
        and Jia Wang

\IEEEcompsocitemizethanks{\IEEEcompsocthanksitem Junwei Yin, Min Gao, Yinqiu Huang, and Jia Wang are with Chongqing University, Chongqing 400044, China. E-mail: \{junweiyin, gaomin, yinqiu, jiawang\}@cqu.edu.cn.
\IEEEcompsocthanksitem Kai Shu is with Illinois Institute of Technology, Chicago, IL 60616, USA. E-mail: kshu@iit.edu.
\IEEEcompsocthanksitem Zehua Zhao is with the University of Sydney, Sydney, NSW 2006, Australia. E-mail: zehua.zhao@sydney.edu.au.
}
\thanks{Manuscript received June 11, 2023; revised xxx, 2023.\\
(Corresponding author: Min Gao.)}
}

%
%

\markboth{Journal of \LaTeX\ Class Files,~Vol.~14, No.~8, August~2015}%
{Shell \MakeLowercase{\textit{et al.}}: Bare Advanced Demo of IEEEtran.cls for IEEE Computer Society Journals}
%



\IEEEtitleabstractindextext{%
\begin{abstract}
The wide dissemination of fake news has affected our lives in many aspects, making fake news detection important and attracting increasing attention.
Existing approaches make substantial contributions in this field by modeling news from a single-modal or multi-modal perspective. 
However, these modal-based methods can result in sub-optimal outcomes as they ignore reader behaviors in news consumption and authenticity verification. For instance, they haven’t taken into consideration the component-by-component reading process: from the headline, images, comments, to the body, which is essential for modeling news with more granularity.
To this end, we propose an approach of \underline{Em}ulating the \underline{be}haviors of \underline{r}eaders (Ember) for fake news detection on social media, incorporating readers’ reading and verificating process to model news from the component perspective thoroughly. 
Specifically, we first construct intra-component feature extractors to emulate the behaviors of semantic analyzing on each component. Then, we design a module that comprises inter-component feature extractors and a sequence-based aggregator. This module mimics the process of verifying the correlation between components and the overall reading and verification sequence. Thus, Ember can handle the news with various components by emulating corresponding sequences. We conduct extensive experiments on nine real-world datasets, and the results demonstrate the superiority of Ember.

\end{abstract}

}

\maketitle

\IEEEdisplaynontitleabstractindextext

%
\IEEEpeerreviewmaketitle

\ifCLASSOPTIONcompsoc
\IEEEraisesectionheading{\section{Introduction}\label{sec:introduction}}
\else
\section{Introduction}
\label{sec:introduction}
\fi

\IEEEPARstart{W}{ith} the rapid development of the Internet, social media has become a crucial part of people’s daily life to share information, exchange opinions, and acquire knowledge. In contrast to traditional media such as newspapers and TV, the propagation process of information on social media has dramatically accelerated. Meanwhile, ensuring the authenticity of information on social media poses a substantial challenge. This is due to every user can express their opinions and share posts they find interesting or agree with. Consequently, social media has become an ideal environment for fake news dissemination. Since the wide dissemination of fake news has been shown to cause serious social consequences \cite{Influ1, Influ2, Influ3}, it is necessary to eliminate fake news from social media to purify the information on social media. However, the ever-increasing number of news on social media makes manual detection a time-consuming and labor-intensive problem. 

To ensure automatic fake news detection, recent research has made significant contributions to detecting fake news from the single-modal or multi-modal perspective. Single-modal methods only use one modal to detect fake news, e.g., text \cite{WWW2019, dEFEND} or images \cite{qimdom}. However, it is challenging to ascertain the authenticity of news by only using text or images \cite{2023IPMProgressive} because news on social media generally contains text and images \cite{2021MM, 2021IPM, HMCAN, CAFE}. In this scenario, multi-modal fake news detection approaches have been proposed to integrate detect-favorable features that come from intra-modal and inter-modal effectively. Some approaches aim at exploring the similarity between news text and images from the semantic perspective \cite{EANN, SAFE}. Others further explore to identify the tampered image based on the text-image similarity \cite{2021IPM, 2023IPMProgressive}. Moreover, some studies extract visual entities to enhance textual and visual representation \cite{2021MM} to boost fake news detection performance.

\begin{figure*}[htbp]
    \centering
    \includegraphics[width=0.9\linewidth,,height=0.3\textwidth]{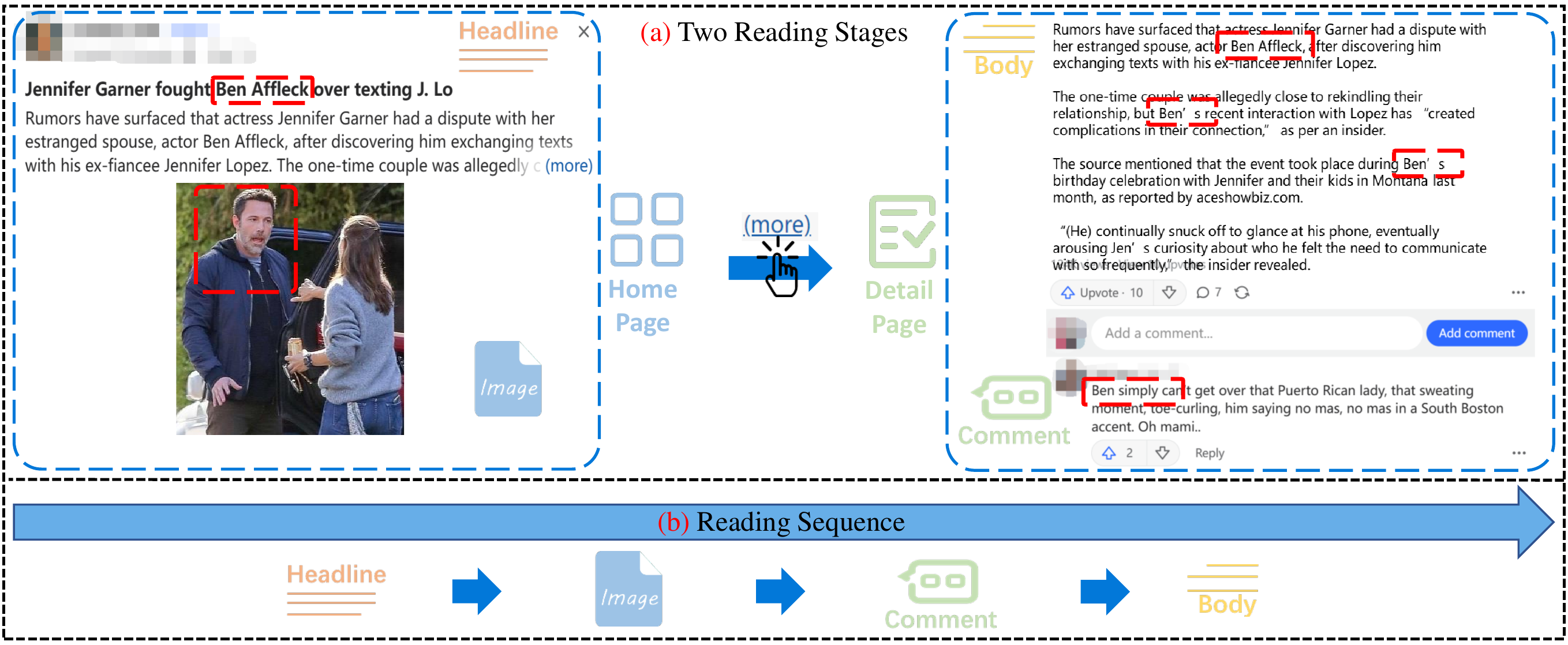}
    \caption{Illustration of reading stages and reading sequence among multiple components, with a news example with four components, i.e., headline, image, comment, and body text. (a) two reading stages from the home page to the detail page. (b) the reading sequence that readers generally prefer among the components.}
    \label{fig:model}
\end{figure*}
Existing fake news detection methods have achieved commendable results, yet they fail to delve deeply into the behaviors exhibited by users when reading news on social media. In general, readers treat news as distinct components rather than as modalities. The components of a news item include headlines, images, comments, bodies, etc., as shown in Fig. 1. Moreover, readers tend to follow a relatively set sequence when reading through these various news components. Consequently, when readers seek to reinforce comprehension or assess authenticity, they often need to compare and confirm the information across different components of the news.
As the sequence of reading reflects the importance of the components and the features between the components (such as semantic differences) aid in reinforcing semantics, both aspects play a crucial role in assessing the authenticity of the news.
However, existing approaches overlook the disparity and hidden clues of inter-component that readers usually leverage to identify news authenticity.

To this end, we propose an approach of Emulating the Behaviors of Readers (Ember) for fake news detection on social media, inspired by the readers' reading behaviors on social media, to take full advantage of inter-component features. In Ember, we recognize that real news components often have high affinity and mutual support, in line with previous studies findings on modal similarity or consistency \cite{SAFE, AAAIHB, 2021IPM}. We utilize this understanding in verification, as illustrated by all components in Fig. 1(a) mentioning the actor $\textit{Ben Affleck}$. Therefore, Ember comprehensively investigates a wide range of news components like readers generally do to model news thoroughly, i.e., refining the process of fake news detection from inter-modality to inter-component. This refinement faces an inevitable challenge: how to excavate the inter-component features simultaneously among different news components and effectively integrate them to promote fake news detection. Existing methods, e.g., \cite{KumarHeadline, AAAIHB, 2021MM, 2023IPMProgressive, HMCAN, SERN} cannot directly tackle the challenge because they elaborate different fusion modules to deal with corresponding inter-component features while ignoring hidden features among the components. 
To tackle this challenge, we first conduct a comprehensive investigation into how readers generally read and verify news on social media and give a definition of the reading and verification sequence. Then, we implement our Ember based on the aforementioned investigation.

In general, due to the layout of social media shown in Fig. 1(a), the process of a reader getting news components can be illustrated as two stages: \textbf{(1)} news headlines and images on the home page of social media; \textbf{(2)} the corresponding body text and readers' comments in the detail page after the reader \textbf{clicks} a piece of news. Readers generally combine the headline and image in stage (1) to assess news authenticity preliminarily, and incorporate the body text and readers’ comments for further verification in stage (2) \cite{ACLFindings21, ICMR2020, eyetrack}. Besides, they generally read shortcut information first and make decisions based on them, e.g., forward news posts by headlines \cite{AAAIHB, TCSSRed}. Since comments are usually concise as the headline, the reading and verification behaviors of readers can be described as a sequence from shortcut information to long body text: headlines, images, comments, and body text (as shown in Fig. 1(b)).

Based on the aforementioned observations, we construct Ember with intra-component feature extractors, an inter-component feature serialization module, and a fake news detector. One intra-component or inter-component feature extractor works for semantic analysis on one news component or verification on two components, respectively; four intra-component feature extractors are used to handle headlines, images, comments, and body text, respectively. Specifically, the intra-component feature extractors aim to emulate the readers' reading process from the local area to the whole component. The inter-component feature serialization module uses inter-component feature extractors with a refinement loss. This aims to emulate reader behaviors of understanding and verifying every two components. A sequence-based aggregator is also used to mimic the entire verification sequence, reflecting the two stages mentioned above. Last, the detector makes a detection based on the features learned by the sequence-based aggregator.

Our main contributions can be summarized as follows: 
\begin{itemize}
    \item{To the best of our knowledge, it is the first exploration for replicating the process of people reading and verifying news on social media in fake news detection from the component perspective.}
    \item{We propose a multi-component fusion method in emulating reader behaviors, which is more suitable and comprehensive for fake news detection in complex scenes than existing methods since components are more fine-grained than modals.}
    \item{We explore the effectiveness of considering the relationship among the components, headlines, images, comments, and body text, in fake news detection.} 
    \item{Extensive experiments on nine real-world datasets illustrate the effectiveness of reading-behavior emulation for fake news detection and the benefits of inter-component feature serialization}
\end{itemize}

\section{Related work}
Given the increasing necessity and attention to fake news detection, many researchers have made significant contributions to this field. Some work \cite{ijcai2020,GCAN,socailBot22} incorporate users’ profiles to assist detection or introduce propagation structures \cite{GACL,EBGCN,22uncertain, RDCL, TBDrumor} to boost fake news detection. As user profiles are related to privacy and most structure-based methods only consider textual information in addition to structure information, we mainly focus on detecting fake news based on single-modal and multi-modal.
\subsection{Single-Modal Methods}
News on social media is rich in various information such as text, images, and so on. The single-modal methods mainly leverage textual or visual features to detect fake news.

\textit{Textual features}. Traditional single-modal methods detect fake news by manually analyzing the writing style of news on social media, such as lexical features, syntactic features, and topic features. \citet{KwonEmotion} propose a periodic time series model to extract the words with strong emotions to distinguish real and fake news, e.g., positive and negative words. Besides, \citet{Rashkin} summarize that first or second-person pronouns and exaggerated words are generally used more in fake news writing. As deep learning methods have shown great potential in fake news detection, some researchers construct deep neural networks (DNN) to model news automatically. \citet{HAN} build a two-layer DNN with the attention mechanism to excavate the significant words and the context information to improve fake news detection performance.
Moreover, \citet{DAFD} propose a domain-adaptive fake news detection framework, which adopts domain-adaptive and adversarial learning on news body text to facilitate fake news detection in different domains. \citet{Kumar2022b} propose a BERT-based model to excavate the long-term dependencies among words of sentences in body text. While \citet{KumarHeadline} and \citet{AAAIHB} further explore the incongruent between news headlines and body text to solve the fake news detection problem. In addition, \citet{TBDmemory} construct an end-to-end memory network to model news from the body text perspective. This network integrates an information retrieval system, which is more effective and accurate.

\textit{Visual features}. \citet{qimdom} point out that news images in fake news always show low resolution and strong emotional information. They propose to excavate the images’ features of the frequency and pixel domain to detect fake news. Besides, \citet{JinImage} propose detecting fake news using images’ visual and statistical features. Specifically, images of real news generally are diverse and high-resolution.

\subsection{Multi-Modal Methods}
Since fake news can be elaborated from multi-perspectives by malicious fake news creators, such as image tampering and body text manipulating \cite{2021IPM}, single-modal may lead to a sub-optimal result. To alleviate this issue, some works propose detecting fake news by using multi-modal data. \citet{HMCAN} construct a hierarchical attention network, which leverages the pre-trained BERT \cite{BERT} and ResNet \cite{ResNet50} to mine textual and visual features, respectively. They further design some context-aware transformers to fuse the inter-modal features. Some works try to detect fake news by categorizing the similarity between text and images \cite{SAFE,2021IPM}. \citet{SAFE} propose a similar capture network to calculate similarity, and news with low similarity is generally fake. While \citet{2021IPM} utilize the pre-trained BERT and ResNet50 network to vectorize textual and visual features. In addition to similarity, \citet{2021IPM,ACLFindings21,2023IPMProgressive} also consider the physical features of news images to assist in identifying fake news. Specifically, \citet{2021IPM} incorporate the error level analysis (ELA) algorithm to judge whether news images have been tampered with. While \citet{2023IPMProgressive} construct a progressive network to model multi-modal news, and \citet{ACLFindings21} propose a multi-modal co-attention network to fuse body textual and visual features, both of them transfer the images into the frequency domain to identify its authenticity. Besides, some works try to improve fake news detection by refining the news features from some specific perspectives; for example, \citet{song2021multimodal} propose a crossmodal attention residual and multichannel convolutional neural networks to decrease the noise from different modals, and \citet{2021MM} design some entity features extractors to extract entities from body text and images, which can be leveraged to enhance news representation. Furthermore, \citet{CAFE} construct a fake news detection model to eliminate fake news by evaluating the cross-modal ambiguity between different modalities.

\subsection{Differences with Existing Models}
As news components are supportive of each other, excavating the inter-component features to detect fake news is meaningful. For single-modal methods, \citet{KumarHeadline} and \citet{AAAIHB} explore the incongruent between news headlines and body text, while \citet{dEFEND} utilize body text and readers’ comments to realize explainable fake news detection. For multi-modal methods, most of them focus on finding the critical detect-favorable features hidden between body text and images \cite{HMCAN,2021MM, BCMF}. \citet{SAFE} further utilize headlines and body text to excavate hidden features between headline\&image and between body text\&image separately. However, this may leave out the hidden features between headlines and body text. Different from the methods above, Ember can thoroughly unearth the inter-component features and model news more comprehensively; this is powered by the fact that Ember first excavates features between every two components, and then integrate these inter-component features according to readers' behaviors. Specifically, for news with three components, Ember considers three combinations, while \citet{SAFE} can only handle two combinations. Moreover, this gap can be exacerbated when news components increase.

\begin{figure*}[htbp]
    \centering
    \includegraphics[width=0.9\linewidth,height=0.5\textwidth]{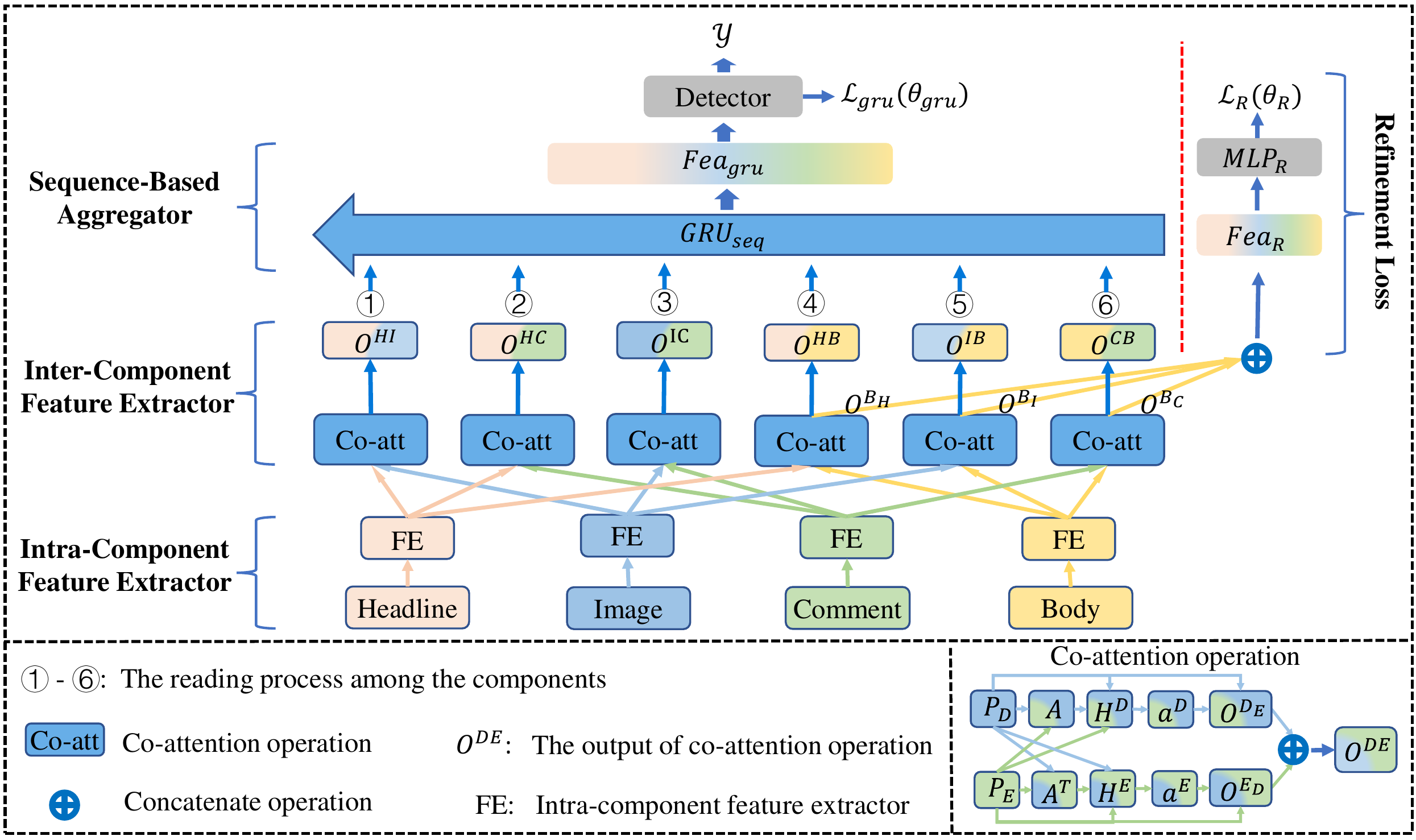}
    \caption{The overview of our proposed Ember model.}
    \label{fig:model}
\end{figure*}

\section{Approach}
\label{sec:format}

Ember detects fake news by emulating readers' reading and verification process, which can effectively extract and integrate the intra-component and inter-component features. As shown in Fig. 2, when considering four news components, Ember comprises four intra-component feature extractors, one inter-component feature serialization module, and one fake news detector. In detail, their functions correspond to the following three: (1) the intra-component feature extractors excavate the semantic features of news components like readers reading on a single component, (2) the inter-component feature serialization module emulates comprehension and verification behaviors among components, and integrate these behaviors according to the reading process, (3) the fake news detector makes a prediction based on the output of the serialization module.


\subsection{Problem Formulation}
To model news comprehensively and multifacetedly, we divide a piece of news into different components. In this paper, we consider four accessible news-related components on social media to validate the effectiveness of Ember, i.e., three textual information (headlines, comments, and body text) and one visual information (news-related images). 

Following previous works \cite{2021MM, DAFD, HMCAN, CAFE}, we regard multi-component fake news detection as a binary classification task. Formally, a piece of news with the headline ($H$), images ($I$), comments ($C$), and body text ($B$) can be written as $N=(H, I, C, B)$. We aim to find a detector $f$ that can explore and integrate the intrinsic information between each two of those components to classify news as true ($y=1$) or fake ($y=0$), i.e., $f(H, I, C, B) \to y \in \{0, 1\}$.

\subsection{Intra-component Feature Extraction}
In this paper, we use four intra-component \textbf{F}eature \textbf{E}xtractors to excavate the semantic features of news \textbf{H}eadlines, \textbf{I}mages, \textbf{C}omments, and \textbf{B}ody text, written as \textbf{HFE}, \textbf{IFE}, \textbf{CFE}, and \textbf{BFE}, respectively. Specifically, each intra-component feature extractor consists of an attention layer and one or two recurrent neural network layers. 

\subsubsection{Textual Feature Extraction}
Since all text information is composed of words and sentences, we naturally formalize the headlines, comments, and body text as $\{H\}_{i=1}^I,  \{C\}_{j=1}^J $, and $\{B\}_{k=1}^K$, respectively, where $I, J$, and $K$ are the number of sentences, and $B_k= \{w_1^k,w_2^k,\cdots,w_{N_k}^k\}$ is a sentence consisting of $N_k$ words. Similarly, $H_i$ and $C_j$ represent a sentence with $N_i$ and $N_j$ words, respectively. The headline is limited to one sentence in formal writing, so setting $I=1$ is more practical and acceptable. Besides, we introduce Glove \cite{glove} to vectorize each word in a sentence into $d$ dimension. 
As many works have shown the salient capability of hierarchical structure to extract features \cite{2021IPM,dEFEND,HMCAN,HAN}, we construct hierarchical attention networks to capture the intra-component semantic information from words to sentences, i.e., HFE, CFE, and BFE, \textit{just like people reading news text from local words to the whole text component on social media}.

Taking the BFE module as an example, to avoid information loss when long sentences are input, two bidirectional Gate Recurrent Unit (Bi-GRU) \cite{GRU} layers are used to encode the words and sentences, respectively. The $\textit{first}$ Bi-GRU layer works for word encoding and can be described as forward and backward reading:

\begin{equation}
    h_b^k = \text{Bi-GRU}_1(w_b^k), b \in \{1,\cdots, N_k\},
\end{equation}
where $k$ represents the $k$-th sentence, and $h_b^k$ is the output of Bi-GRU around word $w_b^k \in \mathbb{R}^{d}$ in two directions.

As the attention mechanism has shown great ability in natural language processing \cite{Attention, BERT, 2021MM, HMCAN}, we further introduce an attention layer to reassign the weight of words according to their importance. The weight of each word can be calculated as follows:

\begin{equation}
    u_b^i = \text{tanh}(W_b^{k}h_b^k+b_b^k),
\end{equation}
\begin{equation}
  \alpha_b^k=  \frac{\text{exp}(u_b^k U_b^T)}{\Sigma_{n=1}^{N_k}\text{exp}(u_n^k U_b^T)},
\end{equation}
where $U_b$, $W_b^k$, and $b_b^k$ are learnable parameters of the network, and $\alpha_b^k$ is the attention weight of word $w_b^k$. $u_b^k$ is the implicit representation of $h_b^k$ obtained by a fully connected layer with $tanh$ activation function. Therefore, a weighted sentence representation vector $s^k \in \mathbb{R}^{2d\times1}$ is calculated by
\begin{equation}
    s^k= \Sigma_{b=1}^{N_k} \alpha_b^k h_b^k.
\end{equation}

To further excavate the context information of sentences, we introduce the $\textit{second}$ Bi-GRU layer after the attention layer, which imitates the process of the user reading news from sentence to sentence. Therefore, the feature of $k$-th sentence can be represented as follows, 
\begin{equation}
    h^k= \text{Bi-GRU}_2(s^k), k \in \{1,\cdots,K\}, 
\end{equation}
where $h^k \in \mathbb{R}^{2d \times 1}$ is the output of Bi-GRU around sentence $s^i$ in two directions.

The structures of both TFE and CFE are similar to BFE and can be regarded as two variants of BFE. Specifically, due to the highly condensed information in the headline, TFE replaces Bi-GRU with bidirectional Long Short Term Memory (Bi-LSTM) as LSTM contains more parameters. While CFE discards the first Bi-GRU layer of BFE since hierarchical architecture is unsuitable for short text \cite{dEFEND}, and comments are less inseparable from news content than headlines in most cases.

\subsubsection{Visual Feature Extraction}
Previous works \cite{HMCAN, qimdom, 2021MM, 2023IPMProgressive} proved that images could supply rich additional information for fake news detection. In this section, we elaborate on the realization of IFE, which emulates the reading process on images and aims to capture the image features from semantic and physical perspectives. 

Specifically, to dig out the physical features of images and perceive whether an image has been tampered with, the error level analysis (ELA) algorithm used in \cite{2021IPM} is incorporated. The ELA algorithm can highlight the malicious stitching parts of fake images to various degrees by setting different error levels \cite{2021IPM}. Following \cite{2021IPM}, we set the error level $r$ as 0.3. Hence, the new image generated by ELA can be represented as:
\begin{equation}
    V_l^{ela}= ELA(V_l, r).
\end{equation}

We then utilize the pre-trained ResNet50 network to vectorize the original and ELA images, 
\begin{equation}
R^{V_l} = ResNet50(V_l), l \in \{1, \cdots, L\},
\end{equation}
\begin{equation}
R^{V_l^{ela}} = ResNet50(V_l^{ela}), l \in \{1, \cdots, L\},
\end{equation}
where $R^{V_l}$ and $R^{V_l^{ela}}$ are the representation of the $l$-th image ($V_l$) in the news and its corresponding ELA image, respectively, and $L$ represents the number of images.

We concatenate $R^{V^{ela}_l}$ and $R^{V_l}$ as $R^{V_l^f}$ so that model can comprehensively deal with image features from semantic and physical perspectives.
Like the text feature extract modules, we leverage an attention layer that takes $R^{V_l^f}$ as input to identify the significance of different parts in these two images and reassign weight to them. Furthermore, to align with text features, we feed the weighted image vector $I^{V_l^{att}}$ into Bi-GRU to attain the final embedding of images,
\begin{equation}
    I^{V_l^{att}} = \text{Attention}(R^{V_l^f}),
\end{equation}
\begin{equation}
    I^{V_l^f} =\text{Bi-GRU}_{img}(I^{V_l^{att}}),
\end{equation}
where $\text{Attention}$ is an abbreviation for the attention mechanism, and its detailed process can refer to Eqs. (2)-(4). $I^{V_l^f} \in \mathbb{R}^{2d \times 1}$ indicates the final representation of image $V_l$.

\subsection{Inter-Component Feature Serialization}
This module emulates readers' comprehension and verification process, comprising six inter-component feature extractors and a sequence-based aggregator. As Fig. 1 illustrates, different news components on social media should complement and support each other, and that’s also one of the basis that people use to distinguish fake news more effectively \cite{PACMHCI21, 2012CSCW}. Besides, studies demonstrate that digging out information from two components \cite{dEFEND, KumarHeadline, AAAIHB} or modals \cite{SERN,2021MM,2023IPMProgressive} is effective.
Thus, to replicate the readers' verification process on news components, we combine every two of these components and then utilize inter-component feature extractors to unearth the affinity of each combination, respectively. In addition, we design a refinement loss to guide the global news representation learning process. Afterward, we concatenate the inter-component information as a crafted sequence and feed it into the sequence-based aggregator. In this way, we can capture the discriminative features around the news piece more comprehensively.

\subsubsection{Inter-Component Feature Extraction}
For the modeling process, we adopt the parallel co-attention method proposed by \cite{lu-etal-hierarchical} to realize inter-component feature extractors and emulate the verification process on each combination based on their affinity, e.g., the combination of news headlines and images. The co-attention operation assigns weights according to the affinity of two inputs to each other, where higher affinity means higher authenticity for readers. In detail, the specific co-attention steps of two inputs are illustrated in Eqs. (11)-(15).
We first calculate the affinity matrix $A \in \mathbb{R}^{Q \times N}$ to measure the affinity between the two inputs:
\begin{equation}
    A=tanh(P_E^T W_{m}P_D),
\end{equation}
where $P_D = [P_D^1,P_D^2,\cdots,P_D^N] \in \mathbb{R}^{2d \times N}$ and $ P_E = [P_E^1,P_E^2,\cdots,P_E^Q]  \in \mathbb{R}^{2d \times Q}$ are the two inputs of the co-attention operation, $N$ and $Q$ are the number of sentences or images, and $W_m \in \mathbb{R}^{2d \times 2d}$ is a learnable parameter matrix. Specific to the situation when four components are accessible, $P_D, P_E \in \{ \text{Headline},\text{Image}, \text{Comment}, \text{Body text} \}$, and $P_D \neq P_E $. By invoking $A$, the new representation of $P_D$ and $P_E$ can be calculated via the following formulas,
\begin{equation}
    H^D = tanh(W_D P_D +(W_EP_E) A),
\end{equation}
\begin{equation}
    H^E = tanh(W_EP_E +(W_DP_D) A^T),
\end{equation}
\begin{equation}
    a^D \!=\! softmax(W_{DE}^T H^D), \hspace{0.1cm}
    a^E \!=\! softmax(W_{ED}^T H^E),
\end{equation}
\begin{equation}
    O^{D_E} = \Sigma_{i=1}^N a_i^{D} P_D^i, \hspace{0.4cm}
    O^{E_D} = \Sigma_{j=1}^Q a_j^{E} P_E^j,
\end{equation}
\begin{equation}
    O^{DE} = [O^{D_E},O^{E_D}], 
\end{equation}
where $W_D, W_E \in \mathbb{R}^{k \times 2d}$, and $W_{DE}, W_{ED} \in \mathbb{R}^{1 \times k}$ indicate learnable weight matrix, $a^D \in \mathbb{R}^{1 \times N}$ and $a^E \in \mathbb{R}^{1 \times Q}$ are the attention scores of $P_D$ and $P_E$, respectively. Especially, $O^{D_E} \in \mathbb{R}^{1 \times 2d}$ is the output of $P_D$, which incorporate the weighted information from $P_E$, and vice versa for $O^{E_D}  \in \mathbb{R}^{1 \times 2d}$. Moreover, $O^{DE} \in \mathbb{R}^{1 \times 4d}$ is the final output of the co-attention operation and integrates two news-related components.

Furthermore, analytic thinkers generally read news repeatedly to distinguish fake and real news better, while lazy thinking about news content may lead to poor verification. \cite{congnition}. Hence, we further introduce a classifier to refine the global news representations. Specifically, this refinement process is conducted on the \textit{last component} in the reading sequence (here is the body text component). Because only then can the readers obtain all news components (i.e., the entire news article). To emulate the behaviors of readers revisiting the whole news to enhance their understanding, we concatenate three enhanced body text representations calculated by Eq. (15): $Fea_{R} = [O^{B_H}, O^{B_I}, O^{B_C}]$,
where $O^{B_H}, O^{B_I}$, and $O^{B_C}$ are the body text representations fused news headlines, images, and comments, respectively. Therefore, $Fea_{R}$ is the revisit global news representation that needs to be refined.

To thoughtfully ascertain the authenticity of news by revisiting news content as readers do, we feed $Fea_R$ to a two-layer MLP to predict its label and introduce the cross-entropy function to calculate the refinement loss:
\begin{equation}
    G_{R} = MLP_{R}(Fea_{R}),
\end{equation}
\begin{equation}
    \mathcal{L}_{R}(\theta_{R})=-\sum_{i=1}^{N} [Y_{i} \log \left(G_{R}^{i}\right)+\left(1-Y_{i}\right) \log \left(1-G_{R}^{i}\right)],
\end{equation}
where $\theta_{R}$ represents the parameters of the model, which is optimized by $\mathcal{L}_{R}$, and $N$ indicates the number of news. We use $Y_i$ and $G_{R}^{i}$ to denote the ground-truth label and predicted probability of $i$-th news, respectively.

\subsubsection{Sequence-Based Aggregator}
To model news more comprehensively, we take every two of the four news components as the inputs of the inter-component extractor. In this case, each component will be dealt with $n-1$ times. We eventually concatenate the outputs of inter-component extractors based on the order of people generally reading news on social media, i.e., the reading sequence mentioned in Section 1 and Fig. 1(b). Thus, the preliminary news representation can be formalized as $Fea = [O^{HI}, O^{HC}, O^{IC}, O^{HB}, O^{IB}, O^{CB}]$. 

To reflect the readers' reading and verification sequence, we incorporate a GRU layer (sequence model) as the aggregator to aggregate the features from different combinations. As GRU is plagued by the long-term dependencies problem \cite{LTDP1, LTDP2}, we further utilize the GRU to integrate features in the \textit{backward} direction, i.e., let the more crucial part be the later input to alleviate the long-term dependencies problem from the data perspective.
\begin{equation}
    Fea_{gru} = \overleftarrow{GRU_{seq}}(Fea),
\end{equation}
where $Fea_{gru}$ indicates the final news representation that integrates the features from different news components.

\subsection{Multi-Component Fake News Detector}
To detect multi-component fake news based on the $Fea_{gru}$ learned by $GRU_{seq}$, a two-layer MLP (Multi-Layer Perception) is incorporated to predict the label of the news.
\begin{equation}
    G_{gru} = MLP_{gru}(Fea_{gru}),
\end{equation}
where $G_{gru}$ denotes the predicted probabilities of news, and $MLP_{gru}$ indicates the MLP layer used to detect $Fea_{gru}$'s label. This paper uses cross-entropy as the objective function of the model:
\begin{equation}
    \!\mathcal{L}_{gru}\!(\theta_{gru})\!=\! -\! \sum_{i=1}^{N} [Y_{i}\! \log \! \left(G_{gru}^{i}\right) \!+ \! \left(1 \! - \!Y_{i}\right) \! \log \! \left(1\!-\!G_{gru}^{i}\right)\!],
\end{equation}
where $\theta_{gru}$ denotes the parameters that is optimized by $\mathcal{L}_{gru}$, and $G_{gru}^{i}$ indicates the predicted probability of $i$-th news.

The final object of Ember is to classify a piece of news as fake or true through intra-component semantic information and inter-component affinity. Therefore, we combine Eq.(18) and Eq.(21) as the final loss function to learn a better multi-component news representation, 
\begin{equation}
    \mathcal{L}(\theta^*)= \mathcal{L}_{gru}(\theta_{gru})+\lambda \mathcal{L}_{R}(\theta_{R}),
\end{equation}
where $\theta^*$ represents the optimal parameters that are learned by minimizing $\mathcal{L}(\theta^*)$, and $\lambda$ is a hyperparameter to balance the refinement degree on global news representation.

\begin{table*}[]
    \centering
    \caption{The statistics of the datasets.}
    \resizebox{0.9\linewidth}{!}{
        \begin{tabular}{cccccccccc}
            \hline
            Dataset & Satire & MisCon & ManCon & FalCon & ImpCon & Compre & GossipCop & PolitiFact2 & PolitiFact7\\
            \hline
            \# fake news & 1,499 & 1,500 & 1,500 & 1,500 & 1,499 & 1,500 & 2,505 & 107 & 396\\
            \# real news & 2,999 & 2,999 & 2,999 & 2,999 & 2,999 & 2,999 & 9,302 & 179 & 352\\
            \# commets & 57,119 & 60,038 & 87,701 & 167,376 & 77,819 & 91,285 & 31,023 & 13,979 & 43,487\\
            \# images & 4,498 & 4,499 & 4,499 & 4,499 & 4,498 & 4,499 & 11,807 & 286 & 286\\
            \hline
        \end{tabular}
        }
\end{table*}

\begin{table*}[]
\caption{Performance comparison of different methods on four datasets.}
\resizebox{\linewidth}{!}{
\begin{tabular}{cccccccccccccc}
\hline
Dataset                  & Method       & Type & Acc & Prec & Rec & F1                    & Dataset                  & Method               & Type    & Acc & Prec & Rec  & F1  \\ \hline
\multirow{12}{*}{PolitiFact2} & HAN\_H  &  S(1) & 0.845 & 0.851 & 0.846 & 0.841 & \multirow{12}{*}{PolitiFact7}        & HAN\_H                & S(1)    & 0.812 & 0.822 & 0.816 & 0.813 \\
                            & HAN       &  S(1) & 0.857 & 0.898 & 0.856 & 0.860  &                                     & HAN                   & S(1)    & 0.803 & 0.806 & 0.804 & 0.801  \\
                            & Text\_CNN\_H&S(1) & 0.825 & 0.828 & 0.825 & 0.888  &                                     & Text\_CNN\_H          & S(1)    & 0.839 & 0.846 & 0.838 & 0.837 \\
                            & Text\_CNN &  S(1) & 0.807 & 0.814 & 0.805 & 0.808  &                                     & Text\_CNN             & S(1)    & 0.826 & 0.826 & 0.826 & 0.826 \\
                            & dEFEND\_H &  S(2) & 0.857 & 0.871 & 0.856 & 0.859  &                                     & dEFEND\_H             & S(2)    & 0.838 & 0.843 & 0.838 & 0.838\\
                            & dEFEND    &  S(2) & 0.821 & 0.847 & 0.822 & 0.825 &                                      & dEFEND                & S(2)    & 0.826 & 0.830 & 0.826 & 0.826 \\
                            & HMCAN\_H  &  M(2) & \underline{0.912} & \underline{0.913} & \underline{0.912} & \underline{0.913}  & & HMCAN\_H  & M(2)    & 0.852 & 0.857 & 0.852 & 0.843 \\
                            & HMCAN     &  M(2) & 0.895 & 0.896 & 0.895 & 0.894  &                                     & HMCAN                 & M(2)    & 0.873 & \underline{0.882} & 0.873 & 0.873 \\
                            & CAFE\_H   &  M(2) & 0.828 & 0.833 & 0.827 & 0.826&                                       & CAFE\_H               & M(2)    & \underline{0.880} & \underline{0.882} & \underline{0.880} & \underline{0.879} \\
                            & CAFE      &  M(2) & 0.741 & 0.760 & 0.742 & 0.740 &                                      & CAFE                  & M(2)    & 0.831 & 0.839 & 0.832 & 0.830 \\
                            & SAFE      &  M(3) & 0.828 & 0.720 & 0.857 & 0.783  &                                     & SAFE                  & M(3)    & 0.820 & 0.870 & 0.769 & 0.816 \\
                            & \textbf{Ember}   &  M(4) & \textbf{0.964} & \textbf{0.966} & \textbf{0.964} & \textbf{0.964}  & & \textbf{Ember}     & M(4)    & \textbf{0.932} & \textbf{0.933} & \textbf{0.932} & \textbf{0.932} \\ \cline{2-14}
                            & Improve(\%)    &       & 5.702 & 5.805 & 5.702 & 5.586&                                  &                       &         & 5.909 & 5.782 & 5.909 & 6.030 \\ \hline

\multirow{12}{*}{GossipCop} & HAN\_H & S(1) & 0.828 & 0.806 & 0.828 & 0.801  & \multirow{12}{*}{Compre}                & HAN\_H                & S(1)    & 0.714 & 0.693 & 0.714 & 0.687\\
                            & HAN    & S(1) & 0.847 & 0.833 & 0.847 & 0.828 &                                          & HAN                   & S(1)     & -- & -- & -- & -- \\
                            & Text\_CNN\_H & S(1) & 0.827 & 0.806 & 0.827 & 0.794 &                                    & Text\_CNN\_H          & S(1)     & 0.760 & 0.749 & 0.759 & 0.744 \\
                            & Text\_CNN & S(1) & 0.812 & 0.788 & 0.812 & 0.750 &                                       & Text\_CNN             & S(1)     & -- & -- & -- & --  \\
                            & dEFEND\_H & S(2) & 0.835 & 0.822 & 0.835 & 0.819  &                                      & dEFEND\_H             & S(2)     & \underline{0.793} & 0.786 & \underline{0.793} & \underline{0.784}\\
                            & dEFEND    & S(2) & \underline{0.870} & 0.860 & \underline{0.871} & \underline{0.861} &   & dEFEND                & S(2)     & -- & -- & -- & -- \\
                            & HMCAN\_H  & M(2) & 0.846 & 0.832 & 0.845 & 0.830  &                                      & HMCAN\_H              & M(2)     & 0.782 & 0.775 & 0.782 & 0.773 \\
                            & HMCAN     & M(2) & 0.844 & 0.830 & 0.844 & 0.832  &                                      & HMCAN                 & M(2)     & -- & -- & -- & --  \\
                            & CAFE\_H   & M(2) & 0.804 & \textbf{0.899} & 0.804 & 0.847 &                              & CAFE\_H               & M(2)     & 0.751 & \underline{0.815} & 0.752 & 0.771 \\
                            & CAFE      & M(2) & 0.753 & 0.848 & 0.753 & 0.795 &                                       & CAFE                  & M(2)     & -- & -- & -- & -- \\
                            & SAFE      & M(3) & 0.810 & 0.572 & 0.462 & 0.511  &                                      & SAFE                  & M(3)     & -- & -- & -- & --  \\
                            & \textbf{Ember} & M(4) & \textbf{0.886} & \underline{0.879} & \textbf{0.886} & \textbf{0.877}  & & \textbf{Ember}  & M(4) & \textbf{0.826} & \textbf{0.823} & \textbf{0.826} & \textbf{0.824} \\ \cline{2-14} 
                            & Improve(\%)  &      & 1.839 & -2.225 & 1.839 & 1.858 &                                         &               &          & 4.161 & 0.982 & 4.161 & 5.102 \\ \hline
\end{tabular}}
\end{table*}

\section{Experiments and results}
\label{sec:pagestyle}

In this section, extensive experiments are conducted on nine real-world datasets to evaluate the effectiveness of Ember in fake news detection. Specifically, we raise four research questions below to guide the experiments.

\textbf{RQ1}: Can Ember outperform other advanced methods in fake news detection?

\textbf{RQ2}: How about the generalization of Ember in detecting different types or domains of fake news, and to what extent the news categories will impact the model’s performance?

\textbf{RQ3}: How effectively does each part of Ember in improving its performance?

\textbf{RQ4}: Can Ember learn high-quality news representation and how effective affinity calculation and refinement loss are?

\subsection{Datasets}
\label{sec:typestyle}
We perform experiments on nine real-world datasets to assess the effectiveness of Ember comprehensively. The details of these nine datasets are as follows:

FakeNewsNet \cite{shu2020fakenewsnet} includes two fake news datasets collected from two fact-checking platforms: GossipCop\footnote{https://www.gossipcop.com/} and PolitiFact\footnote{https://www.politifact.com/}. GossipCop is a dataset in the domain of gossip, while PolitiFact is in politics. Both include news headlines, images, comments, and body text. Since the PolitiFact dataset only owns 286 images, we further construct \textit{two} sub-datasets named PolitiFact2 and PolitiFact7, where PolitiFact2 includes 286 news with 286 images, and PolitiFact7 includes 748 news with 286 images.

Fakeddit \cite{Nakamura2020FakedditAN} is a fake news dataset collected from Reddit\footnote{https://www.reddit.com/} and contains headlines, images, and comments, including five types of fake news:  \textbf{\underline{satire}} (content with a satirical tone), \textbf{\underline{mis}leading \underline{con}tent} (intentionally manipulated content), \textbf{\underline{man}ipulated \underline{con}tent} (manual photo edit), \textbf{\underline{fal}se \underline{con}nection} (mismatch of text and images), and \textbf{\underline{imp}oster \underline{con}tent} (bot-generated content). We construct \textit{six} sub-datasets according to the news category mentioned before, named: Satire, MisCon, ManCon, FalCon, ImpCon, and Compre, where Compre is a \underline{compre}hensive dataset that includes all five types of fake news. 

It is worth noting that GossipCop and PolitiFact are domain-specific datasets but not category-specific, while Staire, MisCon, FalCon, ImpCon, and Compre are category-specific datasets but not domain-specific. Especially, Compre is neither domain-specific nor category-specific. The statistics of these datasets are shown in Table 1.

\subsection{Baselines}
To assess the superiority of Ember, we compare it with six advanced baseline methods as follows, where 'S(n)' and 'M(n)' indicate the single-modal method and multi-modal method used \textit{n} news-related components, respectively. Moreover, the first five methods use body text as textual information, while SAFE \cite{SAFE} additionally considers the news headlines. To further evaluate the effectiveness of headlines in fake news detection, we substitute the body text used in the first five baselines with the headlines, renaming these methods as 'baseline\_H,' e.g., 'HAN\_H.'

S(1) \textbf{HAN} \cite{HAN}: HAN uses body text to identify fake news. This method builds a hierarchical attention network to excavate word-level and sentence-level features.

S(1) \textbf{Text\_CNN} \cite{Text_CNN}: Text\_CNN leverages news body text to model news. They incorporate conventional kernels to capture the features that come from different dimensions.

S(2) \textbf{dEFEND} \cite{dEFEND}: dEFEND proposes an explainable fake news detection model, which utilizes body text and users’ comments to find \textit{k} explainable comments to improve fake news detection.

M(2) \textbf{HMCAN} \cite{HMCAN}: HMCAN utilizes BERT and ResNet50 to extract textual and visual features, which constructs a multi-modal contextual attention network to integrate cross-modal features.

M(2) \textbf{CAFE} \cite{CAFE}: CAFE constructs a cross-modal alignment module to transform different modals’ features into a shared semantic space, then evaluates the ambiguity between different modalities.

M(3) \textbf{SAFE} \cite{SAFE}: SAFE leverages the similarity of text and images to detect fake news. Specifically, they utilize a pre-trained image-to-text model to transform an image into text and then calculate their similarity.

\subsection{Implementation Details}
We implement all the baselines and the proposed Ember using NVIDIA GeForce RTX 3090 and Pytorch 1.9.0. We first use Glove \cite{glove} and ResNet50 \cite{ResNet50} to encode text and images, respectively. The words' dimensions are 100, and the images' dimensions are 1024. The ResNet50 network is pre-trained on the ImageNet set, and its weights are frozen. Besides, the dimension of Bi-GRU is 50, and the dimension of the inter-component extractor's output is 200. The hyperparameters $\lambda$ in Eq.(22) are 0.6, 1, 0.1, and 0.4 on PolitiFact2, PolitiFact7, GossipCop, and Compre respectively. We split every dataset into training, validation, and testing sets with a ratio of 8:1:1. The model is trained in a batch size of 64 and 100 epochs with a learning rate of 1e-3. We incorporate early stopping \cite{earlystop} to interrupt the training process to avoid overfitting when the validation loss stops decreasing by 8 epochs, and Adaptive Moment Estimation (Adam) \cite{Adam} is used as the optimizer.

\begin{table}[]
    \centering
    \caption{Ablation study on PolitiFact7 and GossipCop datasets.}
    \scalebox{1}{
        \begin{tabular}{cccccc}
            \hline
            Dataset & Variants & Acc & Prec & Rec & F1\\
            \hline
            \multirow{9}{*}{PolitiFact7}& Ember/H & 0.851 & 0.859 & 0.851 & 0.851\\
            & Ember/I & 0.905 & 0.914 & 0.905 & 0.905 \\
            & Ember/C & 0.905 & 0.914 & 0.905 & 0.905 \\
            & Ember/B & 0.865 & 0.867 & 0.865 & 0.865\\
            & Ember/ELA & 0.905 & 0.909 & 0.905 & 0.906 \\
            & Ember/GRU & 0.824 & 0.832 & 0.824 & 0.824 \\
            & Ember-Att & 0.878 & 0.879 & 0.878 & 0.878 \\
            & Ember-BiGRU & 0.878 & 0.882 & 0.878 & 0.879 \\
            & Ember & \textbf{0.932} & \textbf{0.933} & \textbf{0.932} & \textbf{0.932}\\
            \hline

            \multirow{9}{*}{GossipCop}& Ember/H & 0.874 & 0.865 & 0.874 & 0.864\\
            & Ember/I & 0.826 & 0.798 & 0.826 & 0.798\\
            & Ember/C & 0.854 & 0.841 & 0.854 & 0.843\\
            & Ember/B & 0.824 & 0.793 & 0.824 & 0.791\\
            & Ember/ELA & 0.883 & 0.876 & 0.883 & 0.874\\
            & Ember/GRU & 0.861 & 0.849 & 0.861 & 0.846\\
            & Ember-Att & 0.873 & 0.865 & 0.873 & 0.867 \\
            & Ember-BiGRU & 0.868 & 0.858 & 0.868 & 0.854 \\
            & Ember & \textbf{0.886} & \textbf{0.879} & \textbf{0.886} & \textbf{0.877}\\
            \hline
        \end{tabular}
    }
\end{table}

\subsection{Performance Comparison (RQ1)}
To answer RQ1, four metrics are used to measure the performance of algorithms: Accuracy, Precision, Recall, and F1 score. The results of different methods are reported in Table 2, where the highest scores are in bold, and the runner-ups are marked with underlines. 
Ember outperforms the other six baseline methods in most cases on four datasets. Specifically, Ember surpasses the sub-optimal results by 4.16\% in accuracy and 5.10\% in F1 score on Compre, and outperforms the sub-optimal results by at least 5.58\% in all metrics on PolitiFact2 and PolitiFact7 sub-datasets, which illustrates the effectiveness of Ember in detecting fake news. In addition, Ember can detect fake news in different domains (politic and gossip) and handle complex scenes that contain various domains and types of fake news (Compre dataset), which shows that Ember is more practicable and general than baselines.

Moreover, all baseline methods show different results when taking headlines or body text as input. As Table 2 illustrates, some methods (HMCAN, Text\_CNN, and CAFE) are more suitable for using news headlines as detection sources. In contrast, some methods (HAN, dEFEND) can achieve better performance by using body text, demonstrating that headlines and body text are helpful for fake news detection. Besides, multi-modal algorithms usually perform better than single-modal algorithms, which shows the necessity of using multi-modal data. 
\begin{figure}[!t]
\centering
\subfloat[Accuracy]{\includegraphics[width=4.3cm]{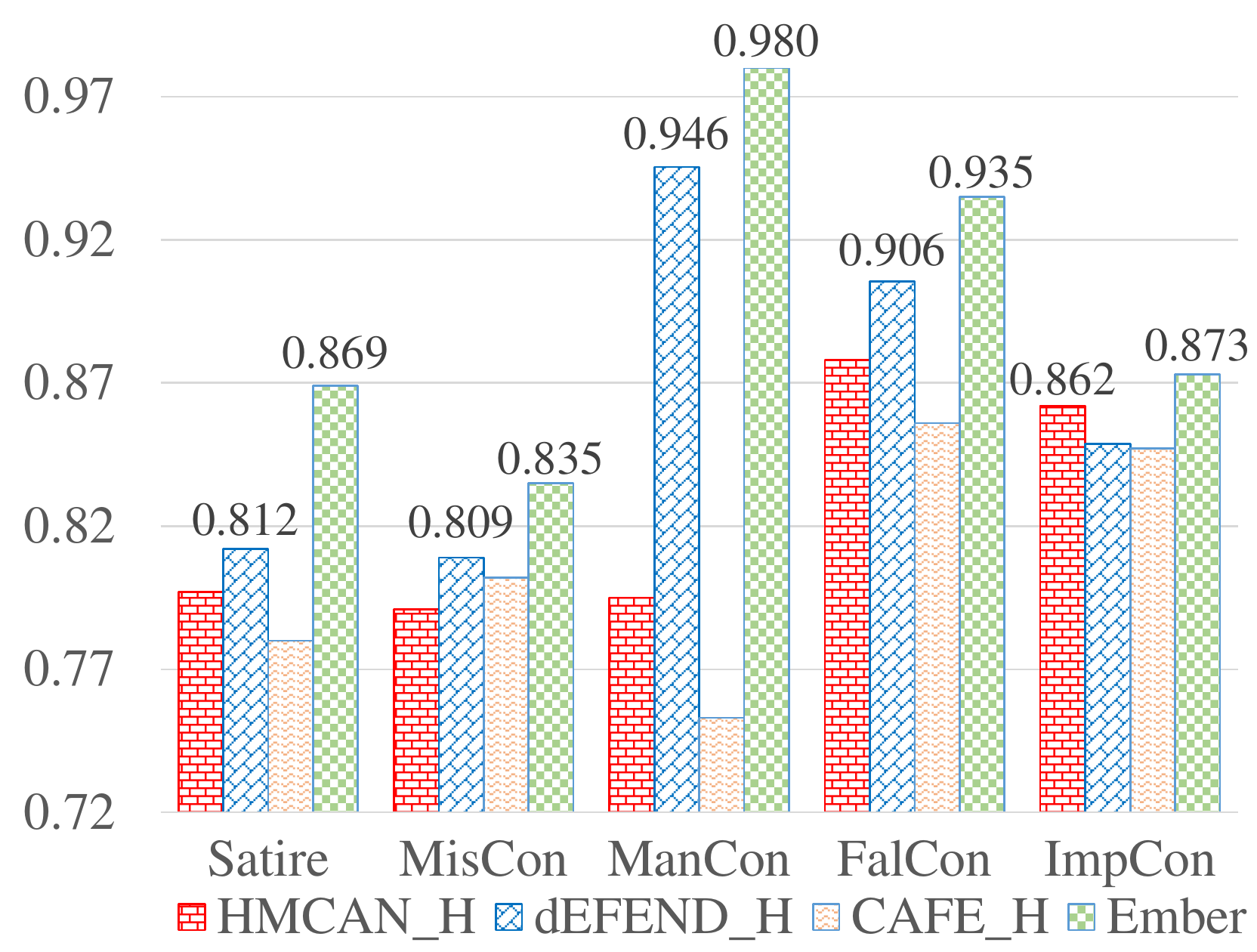}%
\label{fig_first_case}}
\hfil
\subfloat[F1]{\includegraphics[width=4.3cm]{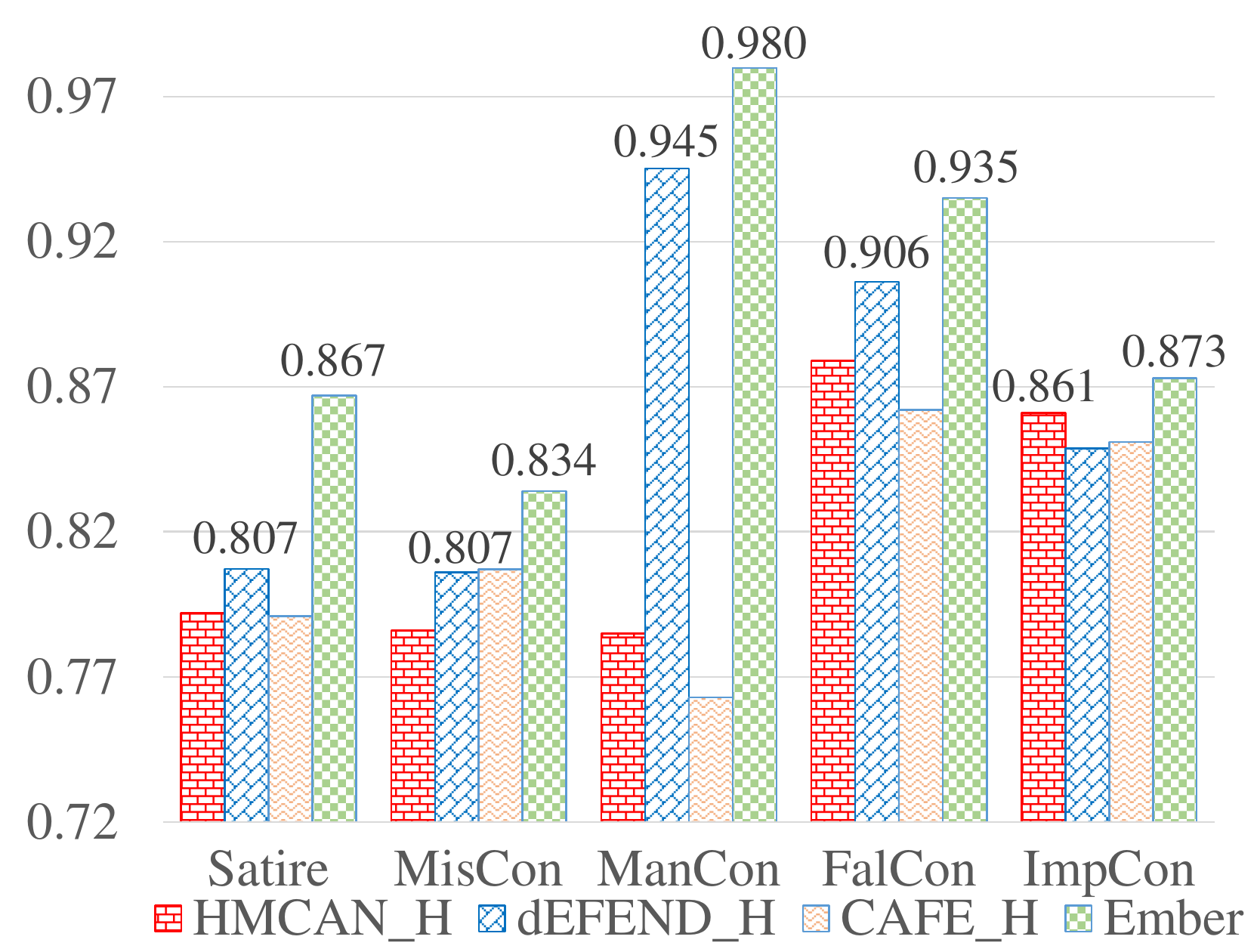}%
\label{fig_second_case}}
\caption{The generalization of Ember on five sub-datasets.}
\label{fig_sim}
\end{figure}

\begin{table}[]
    \centering
    \caption{ The impact of concatenate order on PolitiFact7, GossipCop, and Compre, where HC, HI, etc., are the combinations of two components. $\left[\cdot \right]$ is the concatenate operation, $\left[\cdot \right]$* indicates the order that Ember used.}
    \resizebox{\linewidth}{!}{
        \begin{tabular}{cccccc}
            \hline
            Dataset & Concatenate order & Acc & Prec & Rec & F1\\
            \hline
            \multirow{5}{*}{PolitiFact7}&$\left[(\text{HI}),(\text{HC}),(\text{IC}),(\text{HB}),(\text{IB}),(\text{CB})\right]$* & \textbf{0.932} & \textbf{0.933} & \textbf{0.932} & \textbf{0.932} \\
            &$\left[(\text{HB}),(\text{IB}),(\text{CB}),(\text{HI}),(\text{HC}),(\text{IC})\right]$ & 0.878 & 0.882 & 0.878 & 0.879 \\
            &$\left[(\text{IC}),(\text{HI}),(\text{HC}),(\text{HB}),(\text{IB}),(\text{CB})\right]$ & 0.892 & 0.894 & 0.892 & 0.892 \\
            &$\left[(\text{HI}),(\text{HC}),(\text{IC}),(\text{CB}),(\text{IB}),(\text{HB})\right]$ & 0.892 & 0.894 & 0.892 & 0.892 \\
            &$\left[(\text{HI}),(\text{HC}),(\text{IC}),(\text{IB}),(\text{HB}),(\text{CB})\right]$ & 0.919 & 0.920 & 0.919 & 0.919 \\
            \hline

            \multirow{5}{*}{GossipCop}&$\left[(\text{HI}),(\text{HC}),(\text{IC}),(\text{HB}),(\text{IB}),(\text{CB})\right]$* & \textbf{0.886} & \textbf{0.879} & \textbf{0.886} & \textbf{0.877} \\
            &$\left[(\text{HB}),(\text{IB}),(\text{CB}),(\text{HI}),(\text{HC}),(\text{IC})\right]$ & 0.878 & 0.870 & 0.878 & 0.869 \\
            &$\left[(\text{IC}),(\text{HI}),(\text{HC}),(\text{HB}),(\text{IB}),(\text{CB})\right]$ & 0.871 & 0.863 & 0.871 & 0.864 \\
            &$\left[(\text{HI}),(\text{HC}),(\text{IC}),(\text{CB}),(\text{IB}),(\text{HB})\right]$ & 0.869 & 0.861 & 0.869 & 0.863 \\
            &$\left[(\text{HI}),(\text{HC}),(\text{IC}),(\text{IB}),(\text{HB}),(\text{CB})\right]$ & 0.831 & 0.807 & 0.831 & 0.795 \\
            \hline

            \multirow{3}{*}{Compre}&$\left[(\text{HI}),(\text{HC}),(\text{IC})\right]$* & \textbf{0.826} & \textbf{0.823} & \textbf{0.826} & \textbf{0.824} \\
            &$\left[(\text{HI}),(\text{IC}),(\text{HC})\right]$ & 0.800 & 0.795 & 0.800 & 0.796 \\
            &$\left[(\text{IC}),(\text{HC}),(\text{HI})\right]$ & 0.804 & 0.804 & 0.804 & 0.804 \\
            
            \hline
        \end{tabular}
    }

\end{table}

\subsection{Study on Models Generalization (RQ2)}
To adapt those six sub-datasets we built from the Fakeddit \cite{Nakamura2020FakedditAN} dataset, we reconstruct Ember with three intra-component feature extractors (HFE, CFE, and IFE) and three inter-component feature extractors. Hence, we can evaluate Ember's generalization in detecting various fake news and explore the impact of fake news categories on model detection performance. Since the body text is unavailable in these sub-datasets, we utilize HMCAN\_H, dEFEND\_H, and CAFE\_H as the compare baselines. In addition, for the refining operation, we conduct it on comments ($Fea_R = [O^{C_H}, O^{C_I}]$) since it is the last component in the reading sequence. Only the methods that take at least two kinds of news components as inputs are considered, and SAFE is not feasible on these datasets since it requires body text and headlines simultaneously. The results are represented in Fig. 3, which suggests that Ember outperforms other methods in generalization, i.e., Ember can detect various types of fake news better than baseline methods. Meanwhile, easy to find that the performance of an algorithm varies significantly in detecting different categories of fake news, while results on Compre prove that Ember can model those five types of fake news comprehensively and outperform other baselines. Moreover, these results demonstrate that it's hard to design an algorithm to detect various categories of fake news. So, it's necessary to develop some methods to detect category-specific fake news, and dividing fake news into a specific category can help to explain why it is fake news. We leave this for future work.

\begin{figure}[!t]
\centering
\subfloat[PolitiFact2]{\includegraphics[width=4.3cm]{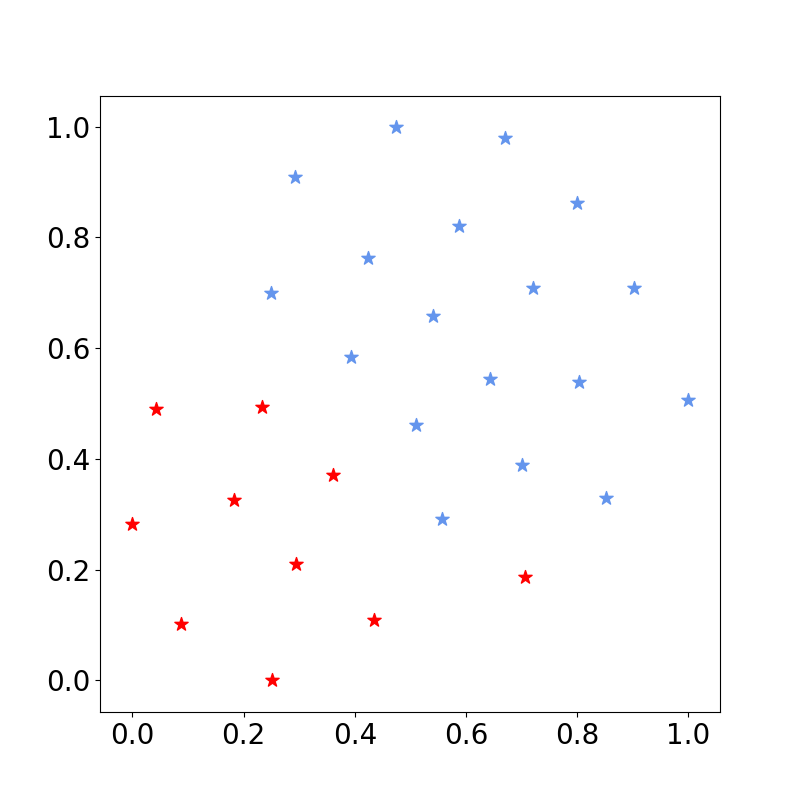}%
\label{fig_first_case}}
\hfil
\subfloat[PolitiFact7]{\includegraphics[width=4.3cm]{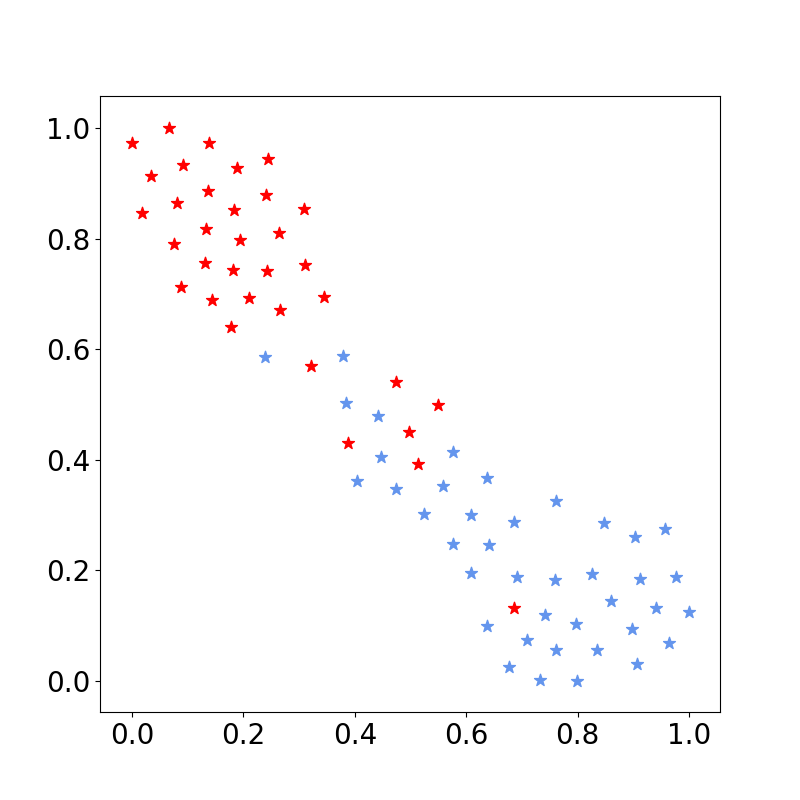}%
\label{fig_second_case}}

\subfloat[GossipCop]{\includegraphics[width=4.3cm]{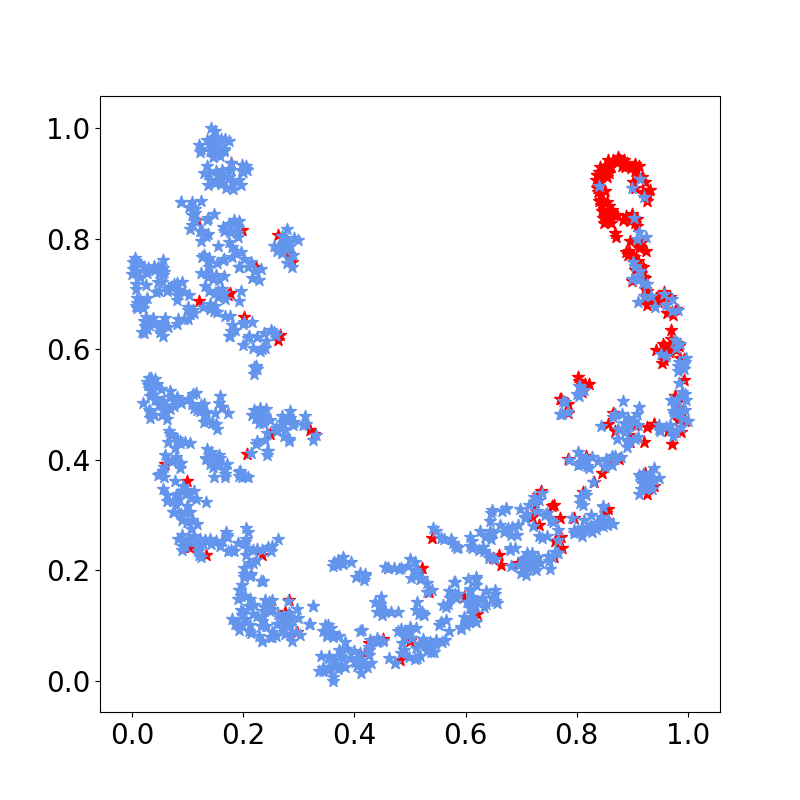}%
\label{fig_third_case}}
\hfil
\subfloat[Compre]{\includegraphics[width=4.3cm]{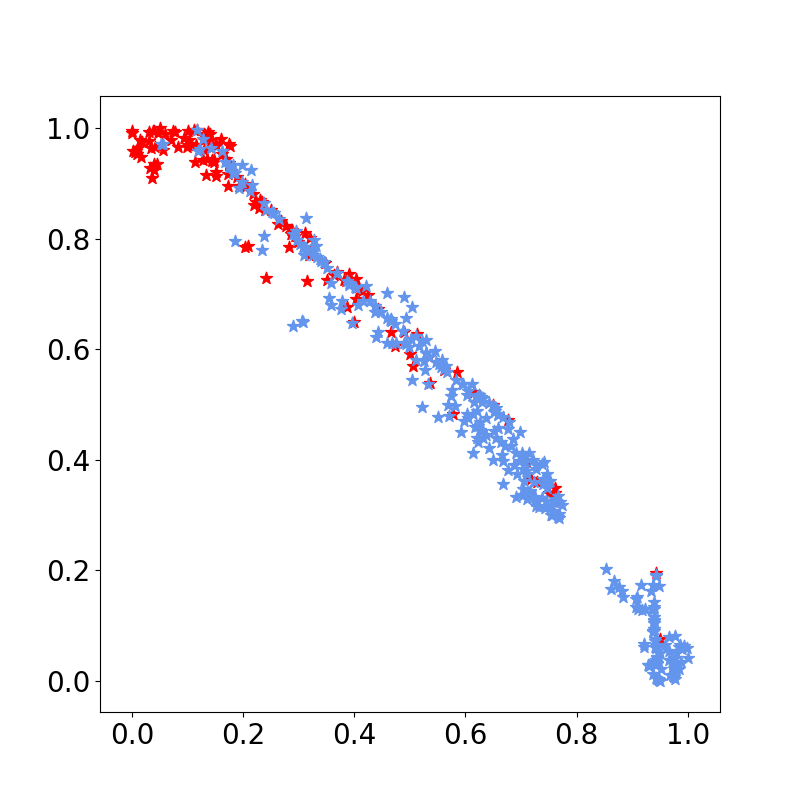}%
\label{fig_forth_case}}
\caption{The generalization of Ember on five sub-datasets.}
\label{fig_vis}
\end{figure}

\subsection{Ablation Study (RQ3)}
To measure the necessity of news headlines, images, comments, and body text and the effectiveness of the inter-component feature serialization module, we compare Ember with eight variants on PolitiFact7 and GossipCop datasets. The detailed results can be reached in Table 3. Ember/H, Ember/I,  Ember/C, and Ember/B are the variants of Ember that are without headlines, images, comments, and body text, respectively. Ember/GRU is the variant of Ember without $GRU_{seq}$ layer, which directly concatenates the extracted features and uses them to detect fake news. Ember-Att and Ember-BiGRU are two variants that use the attention mechanism and Bi-GRU to substitute $GRU_{seq}$ layer, respectively. Ember/ELA is a variant without considering the ELA algorithm. Obviously, any module's absence will lead the sub-optimal results. In detail, for those four components, the headlines and body text contribute the most to fake news detection on PolitiFact7 since Ember/H and Ember/B contribute the worst figures; In contrast, the detect-favorable elements of GossipCop are generally concealed in images and body text. In short, every news-related component is detect-favorable, and mining the hidden information among them is meaningful. 

Besides, the ablation of $GRU_{seq}$ layer decreases Ember's performance on both two datasets, especially on PolitiFact7, which shows readers' reading sequence is beneficial for fake news detection.
Therefore, we further explore the effectiveness of concatenating order, i.e., reading sequence. As Table 4 reported, the concatenated order which stems from readers' reading behavior surpasses others, which proves the rationality and validity of Ember. In general, the headlines and images are the first to be read by readers, while the body text should be the last one due to the layout of social media and readers' general reading sequence.

\begin{table}[]
    \centering
    \caption{Results that discard one combination on three datasets.}
    \scalebox{1}{
        \begin{tabular}{cccccc}
            \hline
            Dataset & Variants & Acc & Prec & Rec & F1\\
            \hline
            \multirow{7}{*}{PolitiFact7} & Ember/HI & 0.905 & 0.908 & 0.905 & 0.905\\
            & Ember/HC & 0.931 & \textbf{0.936} & 0.931 & \textbf{0.933}\\
            & Ember/IC & 0.905 & 0.909 & 0.905 & 0.906\\
            & Ember/HB & 0.865 & 0.870 & 0.865 & 0.865\\
            & Ember/IB & 0.878 & 0.879 & 0.878 & 0.878\\
            & Ember/CB & 0.919 & 0.920 & 0.919 & 0.919\\
            & Ember & \textbf{0.932} & 0.933 & \textbf{0.932} & 0.932\\
            \hline
            \multirow{7}{*}{GossipCop}& Ember/HI & 0.830 & 0.805 & 0.830 & 0.795\\
            & Ember/HC & 0.868 & 0.864 & 0.868 & 0.866\\
            & Ember/IC & 0.872 & 0.863 & 0.872 & 0.864\\
            & Ember/HB & 0.828 & 0.800 & 0.828 & 0.795\\
            & Ember/IB & 0.824 & 0.794 & 0.824 & 0.788\\
            & Ember/CB & 0.827 & 0.799 & 0.827 & 0.794\\
            & Ember & \textbf{0.882} & \textbf{0.874} & \textbf{0.882} & \textbf{0.871}\\
            \hline
            \multirow{4}{*}{Compre}& Ember/HI & 0.811 & 0.807 & 0.811 & 0.808\\
            & Ember/HC & 0.784 & 0.787 & 0.784 & 0.759\\
            & Ember/IC & 0.795 & 0.789 & 0.795 & 0.790\\
            & Ember & \textbf{0.826} & \textbf{0.823} & \textbf{0.826} & \textbf{0.824}\\
            \hline
        \end{tabular}
    }
\end{table}

\begin{figure}[htbp]
    \centering
    \includegraphics[width=\linewidth]{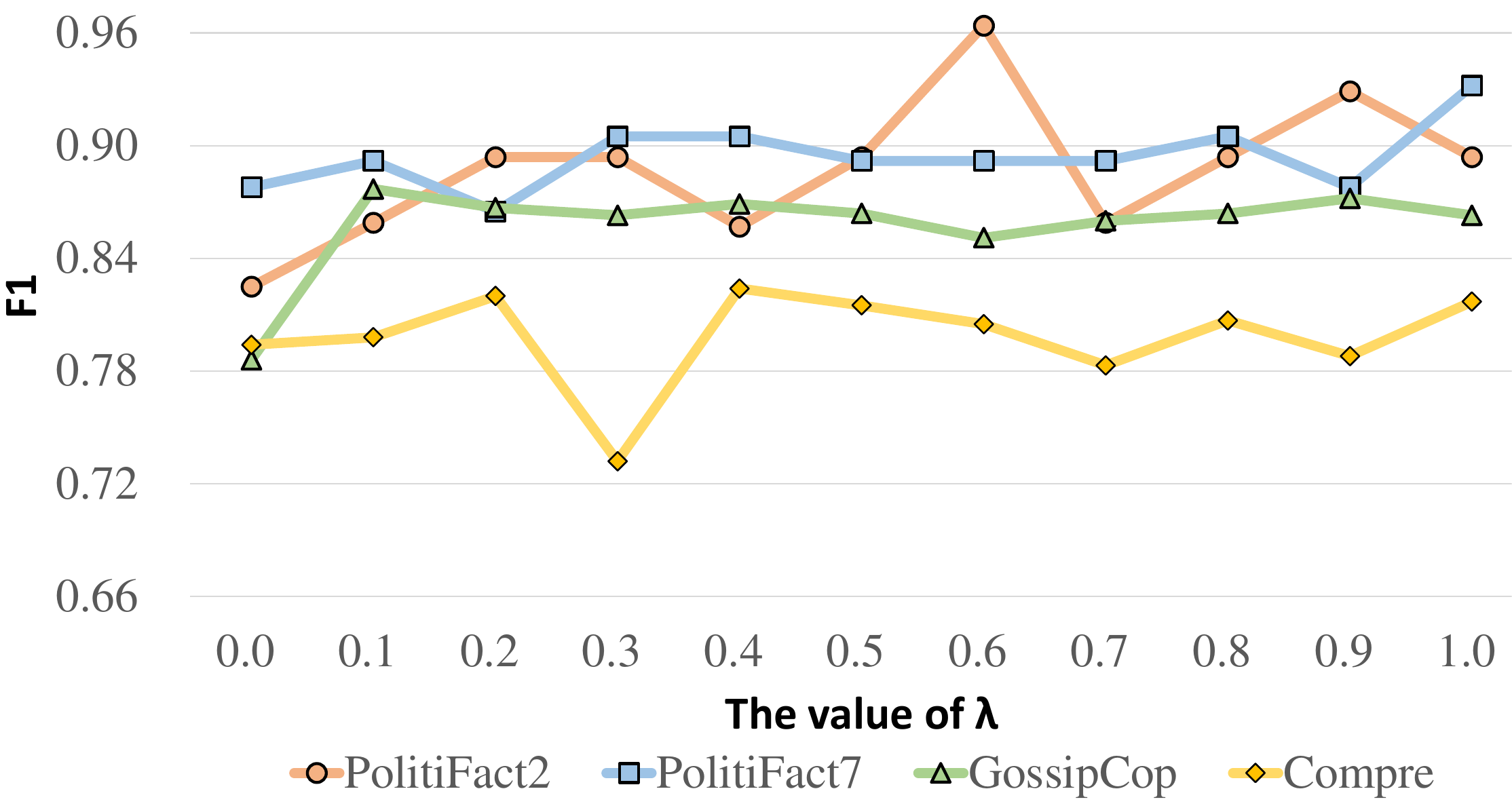}
    \caption{The performance of Ember w.r.t. the values of $\lambda$ on four datasets.}
    \label{fig:HP}
\end{figure}

\subsection{Visualization, Analysis on Combinations, and Hyperparameter (RQ4)}
\textit{Visualization}. We verify that our model can generate high-quality news embeddings by visualizing test samples' embeddings on four datasets. Specifically, we visualize the final news embeddings using the t-Distribution Stochastic Neighbor Embedding (t-SNE) \cite{t-SNE}. As Fig. 4 demonstrated, Ember can generate high-quality news embeddings. Especially in the political domain, fake news and real news own a sharp dividing line.

\textit{Analysis on Combinations}. We further conduct supplementary experiments to justify the contribution of the combination of two components in fake news detection. In detail, we construct some variants of Ember, which discard one combination of those components, e.g., the combination of headlines and images (Ember/HI). The results are shown in Table 5; a performance decrease exists on all datasets when ignoring any combination, demonstrating that digging out the inter-component features between every two components can help detect fake news.

\textit{Hyperparameter}. To evaluate the effectiveness of hyperparameter $\lambda$, we conduct experiments on four datasets, and the results are illustrated in Fig. 5. With the different degrees of refinement, Ember shows various performance improvements (in most cases) in fake news detection, especially on PolitiFact2 and GossipCop, indicating that refinement on news global representation can assist in learning high-quality news representation.

\section{Conclusion}
\label{sec:typestyle}
In this paper, inspired by the readers’ reading and verification behaviors on news components, we innovatively propose a new fake news detection approach called Ember. Ember is more fine-grained, comprehensive, and suitable for multi-modal fake news detection than existing methods, as it models news from the component perspective. 
In detail, we refine the multi-modal fake news detection problem as a multi-component fusion problem, aiming to dig out the intra-component and inter-component features. First, we design intra-component feature extractors to mimic the reading and semantic understanding process on each news component. Then, we construct an inter-component feature serialization module, which is utilized to emulate the comprehension and verification process on every two components and integrate these behaviors as a sequence to reflect the overall reading process. Last, a multi-component fake news detector is constructed to predict news authenticity. Moreover, Ember can adapt to different datasets with various components by adjusting the number of intra-component and inter-component feature extractors.  In short, Ember aims to discover the affinity among various news-related components to model news comprehensively and effectively. Experiments demonstrate the superiority of readers’ behaviors emulation and inter-component feature serialize operation in fake news detection.

\bibliographystyle{IEEEtranN}

\bibliography{refs}


\vspace{-50 mm}
\begin{IEEEbiography}
[{\includegraphics[width=1in,height=1.25in,clip,keepaspectratio]{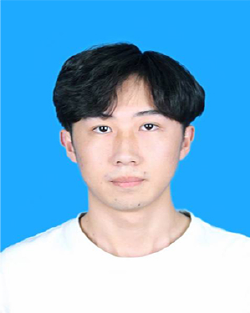}}]{Junwei Yin}
received the B.S. degree in software engineering from Chongqing University, Chongqing, China, in 2022. He is currently pursuing the M.S. degree with the School of Big Data and Software Engineering, Chongqing University, Chongqing, China. 

His research interests include natural language processing and data mining.
\end{IEEEbiography}

\vspace{-40 mm}
\begin{IEEEbiography}
[{\includegraphics[width=1in,height=1.25in,clip,keepaspectratio]{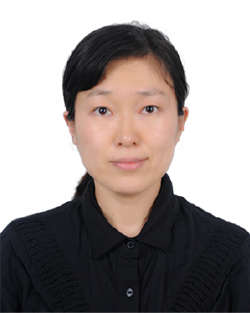}}]{Min Gao} (Member, IEEE) received the M.S. and Ph.D. degrees in computer science from Chongqing University, Chongqing, China, in 2005 and 2010, respectively. She was a Visiting Researcher with the University of Reading, Reading, U.K., and Arizona State University, Tempe, AZ, USA. She is currently a Professor with the School of Big Data and Software Engineering, Chongqing University. She has grants from the National Natural Science Foundation of China, the China Postdoctoral Science Foundation, and the China Fundamental Research Funds for the Central Universities. 

Her research interests include recommendation systems, service computing, and data mining.
\end{IEEEbiography}
\newpage
\enlargethispage{-55mm}
\begin{IEEEbiography}
[{\includegraphics[width=1in,height=1.25in,clip,keepaspectratio]{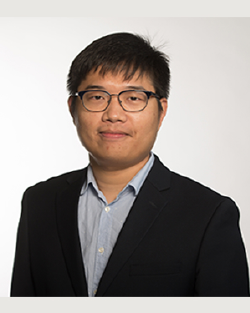}}]{Kai Shu}
received the Ph.D. degree in computer science from Arizona State University, Tempe, AZ, USA, in July 2020. He is a Gladwin Development Chair Assistant Professor with the Department of Computer Science, Illinois Institute of Technology (IIT), Chicago, IL, USA. He was awarded the ASU Fulton Schools of Engineering Dean’s Doctoral Dissertation Award and the CIDSE Doctoral Fellowship 2015 and 2020.

His research interests include data mining, social computing, and their applications in disinformation, education, and healthcare.
\end{IEEEbiography}

\begin{IEEEbiography}
[{\includegraphics[width=1in,height=1.25in,clip,keepaspectratio]{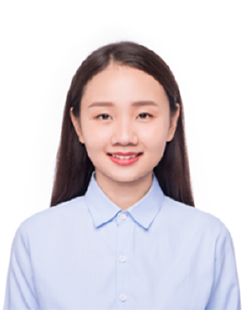}}]{Zehua Zhao}
received the B.S. degree in software engineering from Chongqing University, Chongqing, China, in 2017. She is currently working toward the the Ph.D. degree with the School of Civil Engineering, the University of Sydney, Sydney, Australia. 

Her research interests include intelligent algorithm.
\end{IEEEbiography}

\vspace{3 mm}
\begin{IEEEbiography}
[{\includegraphics[width=1in,height=1.25in,clip,keepaspectratio]{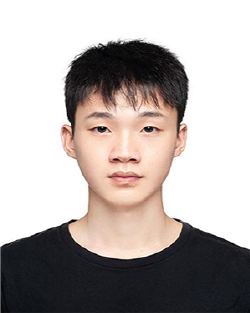}}]{Yinqiu Huang}
received the B.S. degree in network engineering from Southwest Petroleum University, Sichuan, China, in 2021. He is currently pursuing the M.S. degree with the School of Big Data and Software Engineering, Chongqing University, Chongqing, China. 

His research interests include natural language processing and few-shot learning.
\end{IEEEbiography}

\begin{IEEEbiography}
[{\includegraphics[width=1in,height=1.25in,clip,keepaspectratio]{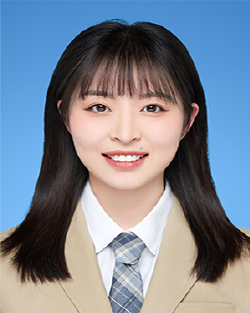}}]{Jia Wang}
received the B.S. degree in software engineering from Chongqing University, Chongqing, China, in 2020. Currently, She is working toward the M.S. degree with the School of Big Data and Software Engineering, Chongqing University, Chongqing, China. 

Her research interests include natural language processing and data mining.
\end{IEEEbiography}







\end{document}